\documentclass{article}
\usepackage{algorithm}
\usepackage{arxiv}

\usepackage[utf8]{inputenc} % allow utf-8 input
\usepackage[T1]{fontenc}    % use 8-bit T1 fonts
\usepackage{hyperref}       % hyperlinks
\usepackage{url}            % simple URL typesetting
\usepackage{booktabs}       % professional-quality tables
\usepackage{amsfonts}       % blackboard math symbols
\usepackage{nicefrac}       % compact symbols for 1/2, etc.
\usepackage{microtype}      % microtypography
\usepackage{graphicx}
\graphicspath{{./images/}}

\title{ZetA: A Hybrid Optimizer Combining Riemann Zeta Scaling with Adam for Robust Deep Learning}

\author{
 Samiksha BC \\
  Department of Computer Science \\
  Indiana University South Bend\\
  samibc@iu.edu
}

\usepackage{amsmath}

\begin{document}
\maketitle

\begin{abstract}
This work introduces \textbf{ZetA}, a novel deep learning optimizer that combines Adam's adaptive gradient updates with dynamic scaling inspired by the Riemann zeta function. To the best of our knowledge, ZetA is the first optimizer to incorporate zeta-based gradient scaling into deep learning optimization. It enhances generalization and robustness through a hybrid update mechanism that integrates adaptive damping, cosine similarity-based momentum boosting, entropy-regularized loss, and Sharpness-Aware Minimization (SAM)-style perturbations. Empirical evaluations on SVHN, CIFAR10, CIFAR100, STL10, and noisy CIFAR10 demonstrate consistent test accuracy improvements over Adam. Experiments employ a lightweight fully connected network trained for five epochs using mixed-precision settings. These results suggest that ZetA is a computationally efficient and robust alternative to Adam, particularly effective in noisy or high-granularity classification tasks.
\end{abstract}

\section{Introduction}

Contemporary deep learning applications require optimizers that balance fast convergence with robustness to noise and overfitting, which are key challenges in real world deployments. Adam, a widely used optimizer known for its adaptive learning rates and momentum-based updates, has become a foundational tool in modern deep learning. However, it often struggles with poor calibration and suboptimal generalization, particularly in noisy or uncertain settings. 

To address these limitations, we introduce \textbf{ZetA}, a novel optimizer that extends Adam with dynamic scaling mechanisms inspired by analytic number theory specifically, the Riemann zeta function. By integrating zeta-based gradient modulation with entropy regularization and sharpness-aware training, ZetA offers improved stability, uncertainty handling, and generalization across a range of challenging datasets.

\section{Related Work}

Optimization in deep learning has seen significant advancements with the emergence of adaptive methods, notably Adam~\cite{kingma2014adam}, which integrates momentum with per-parameter learning rate adaptation, enhancing convergence speed across diverse tasks. Despite its widespread adoption, Adam often converges to sharp minima, leading to suboptimal generalization, especially in noisy or over-parameterized regimes where calibration and robustness are critical. To mitigate these shortcomings, several extensions have been developed: RAdam stabilizes the variance of adaptive learning rates during initial training phases, while AdaBelief~\cite{zhuang2020adabelief} refines second-moment estimation to better align with gradient "beliefs," facilitating convergence toward flatter minima that improve generalization.

Sharpness-Aware Minimization (SAM)~\cite{foret2020sharpness} offers a complementary approach by explicitly targeting low-sharpness solutions in the loss landscape. SAM perturbs weights along the gradient ascent direction before descent, steering networks toward wide, flat minima that enhance generalization and resilience to perturbations. However, these advancements predominantly rely on heuristic or empirically derived scheduling of learning dynamics, lacking a theoretical foundation grounded in mathematical functions with proven convergence properties.

This gap motivates our work, as existing optimizers have yet to explore the potential of functions like the Riemann zeta function for dynamic update scaling. Our proposed ZetA optimizer fills this void by pioneering a hybrid approach, seamlessly blending zeta-based gradient modulation leveraging its analytic convergence behavior—with Adam's adaptive strengths, augmented by entropy regularization and SAM-inspired techniques. This theoretically inspired framework aims to address the limitations of current methods, offering a robust solution for challenging deep learning scenarios.

\section{Method: The ZetA Optimizer}

ZetA is a novel first-order optimizer that augments the Adam optimizer with dynamic scaling inspired by the Riemann zeta function, enhanced by sharpness-aware and entropy-regularized modifications. Its design targets improved generalization, particularly in noisy or low-data regimes, through adaptive scaling, stabilized updates, and robust regularization. The following components define the ZetA optimization algorithm:

\begin{algorithm}

  {\caption{ZetA Optimizer (Hybrid Adam with Riemann Zeta Scaling)}}%

{%
\begin{enumerate}
  \item Input: $\theta$ (parameters), $\eta$ (learning rate), $s_{\min}$, $s_{\max}$, $\beta_1$, $\beta_2$, $\epsilon$, $\rho$ (clip norm), $\delta$ (zeta damp), $\alpha$ (adam mix), $T$ (total steps), $\lambda$ (weight decay), $\gamma$ (SAM rho)
\item Apply gradient clipping: $g_t = \mathrm{clamp}(g_t, -\rho, \rho)$

  \item Compute dynamic zeta exponent: $s_t = s_{\min} + (s_{\max} - s_{\min}) \cdot (1 - |1 - 2 \cdot (t \bmod T)/T|)$
  \item Compute Riemann Zeta: $\zeta(s_t)$ using $\texttt{scipy.special.zeta}(s_t)$

  \item Update EMA and damping: $\delta_t = \delta \cdot (1 + \|g_t\|_\text{EMA} / (1 + \|g_t\|_\text{EMA})) \cdot (1 / \max(0.1, L_\text{EMA})) / (1 - 0.9^t)$

  \item Apply cosine similarity boost: $b_t = 1 + \delta_t \cdot 0.2 \cdot \max(0, \cos(\langle g_t, g_{t-1} \rangle / (\|g_t\| \cdot \|g_{t-1}\| + \epsilon)))$
  \item Apply gradient centralization: $g_t = g_t - \text{mean}(g_t, \text{dim}=1, \text{keepdim}=\text{True})$
  \item Compute moments: $\hat{m}_t = m_t / (1 - \beta_1^t)$, $\hat{v}_t = v_t / (1 - \beta_2^t)$
  \item Compute updates: $u_{\text{adam}} = \hat{m}_t / (\sqrt{\hat{v}_t} + \epsilon)$, $u_{\zeta} = \eta \cdot \hat{m}_t \cdot b_t / (\|g_t\|^{s_t - 1} + \epsilon) \cdot (1 / \zeta(s_t))$
  \item Hybrid update: $u_t = \alpha \cdot u_{\text{adam}} + (1 - \alpha) \cdot u_{\zeta}$
  \item Apply SAM perturbation: $\theta^+ = \theta + \gamma \cdot u_t / (\|u_t\| + \epsilon)$
  \item Update parameters: $\theta_t = \theta_{t-1} - \eta_t \cdot u_t$, where $\eta_t = \eta \cdot (0.5 \cdot (1 + \cos(\pi t / T))) \cdot (1 - \lambda \cdot \eta_t)$
  \item Output: updated parameters $\theta_t$
\end{enumerate}
}
\end{algorithm}

\paragraph{Zeta-Based Gradient Scaling.} Let $g_t$ denote the gradient at iteration $t$. ZetA modulates the learning rate by dynamically scaling the update direction using the Riemann zeta function. The dynamic exponent $s_t$ is computed as:\\

\begin{equation*}
s_t = s_{\min} + (s_{\max} - s_{\min}) \cdot \left(1 - \left|1 - 2 \cdot \frac{t \mod T}{T}\right|\right),
\end{equation*}
\\

where $s_{\min}, s_{\max} \in (1, 2]$, and $T$ is the total number of steps. The zeta function $\zeta(s_t)$ is evaluated using $\texttt{scipy.special.zeta}(s_t)$, ensuring precise computation for $s_t > 1$ (e.g., $\zeta(1.5) \approx 2.612$). The zeta-scaled update component is derived as:\\

\begin{equation*}
u_{\zeta} = \eta \cdot \hat{m}_t \cdot b_t \cdot \frac{1}{\|g_t\|^{s_t - 1} + \epsilon} \cdot \frac{1}{\zeta(s_t)},
\end{equation*}\\

where $b_t$ is the momentum boost, and the term $\frac{1}{\|g_t\|^{s_t - 1} + \epsilon}$ provides adaptive damping based on gradient magnitude, exploiting $\zeta(s_t)$'s decreasing behavior to stabilize large gradients.

\paragraph{Cosine Similarity-Based Momentum Boosting.} To enhance directional stability, ZetA incorporates cosine similarity between successive gradients:\\

\begin{equation*}
\rho_t = \max\left(0, \frac{\langle g_t, g_{t-1} \rangle}{\|g_t\| \cdot \|g_{t-1}\| + \epsilon}\right),
\end{equation*}\\

which is used to compute the boost factor:
\begin{equation*}
b_t = 1 + \delta_t \cdot 0.2 \cdot \rho_t,
\end{equation*}
where $\delta_t$ is the adaptive damping factor. This reinforces consistent descent directions when gradients align, reducing oscillatory updates.

\paragraph{Gradient Centralization.} For weight matrices of shape $[d_{\text{out}}, d_{\text{in}}]$, ZetA applies gradient centralization:
\\

\begin{equation*}
g_t \leftarrow g_t - \text{mean}(g_t, \text{dim}=1, \text{keepdim}=\text{True}),
\end{equation*}
\\

shifting gradients to a zero-mean configuration. This improves generalization and convergence stability, particularly in deep networks~\cite{yong2020gradient}.

\paragraph{Entropy-Regularized Loss.} To mitigate overconfidence, ZetA employs an entropy-regularized loss:
\\

\begin{equation*}
\mathcal{L}_{\text{entropy}} = \mathcal{L}_{\text{CE}} - \lambda \cdot \mathbb{E}_{x} \left[ -\sum_k p_k(x) \log p_k(x) \right],
\end{equation*}\\

where $\mathcal{L}_{\text{CE}}$ is the cross-entropy loss, $p_k(x)$ is the predicted probability for class $k$, and $\lambda = 0.01$ is the regularization coefficient. This promotes smoother posterior distributions, enhancing calibration under uncertainty.

\paragraph{Sharpness-Aware Perturbation.} ZetA integrates a SAM-inspired perturbation to seek flatter minima. The perturbed weights are computed as:\\

\begin{equation*}
\theta^+ = \theta + \gamma \cdot \frac{u_t}{\|u_t\| + \epsilon},
\end{equation*}\\

where $\gamma$ is the SAM rho parameter, and the gradient is evaluated at $\theta^+$ to encourage robustness to small parameter perturbations.

\paragraph{Adaptive Damping and Moment Updates.} The adaptive damping factor $\delta_t$ is updated using an exponential moving average (EMA) of the gradient norm and loss:\\

\begin{equation*}
\|g_t\|_\text{EMA} = 0.9 \cdot \|g_{t-1}\|_\text{EMA} + 0.1 \cdot \|g_t\|,
\end{equation*}\\

\begin{equation*}
L_\text{EMA} = 0.9 \cdot L_{t-1} + 0.1 \cdot \mathcal{L},
\end{equation*}
\\

followed by:
\begin{equation*}
\delta_t = \delta \cdot \left(1 + \frac{\|g_t\|_\text{EMA}}{1 + \|g_t\|_\text{EMA}}\right) \cdot \frac{1}{\max(0.1, L_\text{EMA})} \cdot \frac{1}{1 - 0.9^t}.
\end{equation*}

Moment estimates are updated as:
\begin{equation*}
m_t = \beta_1 \cdot m_{t-1} + (1 - \beta_1) \cdot g_t,
\end{equation*}
\begin{equation*}
v_t = \beta_2 \cdot v_{t-1} + (1 - \beta_2) \cdot g_t^2,
\end{equation*}\\

with bias correction:
\begin{equation*}
\hat{m}_t = \frac{m_t}{1 - \beta_1^t}, \quad \hat{v}_t = \frac{v_t}{1 - \beta_2^t}.
\end{equation*}

\paragraph{Final Update Rule.} The parameter update combines Adam and Zeta components:\\

\begin{equation*}
u_{\text{adam}} = \frac{\hat{m}_t}{\sqrt{\hat{v}_t} + \epsilon},
\end{equation*}
\begin{equation*}
u_{\zeta} = \eta \cdot \hat{m}_t \cdot b_t \cdot \frac{1}{\|g_t\|^{s_t - 1} + \epsilon} \cdot \frac{1}{\zeta(s_t)},
\end{equation*}
\begin{equation*}
u_t = \alpha \cdot u_{\text{adam}} + (1 - \alpha) \cdot u_{\zeta},
\end{equation*}\\

with the learning rate scheduled as:
\begin{equation*}
\eta_t = \eta \cdot \left(0.5 \cdot \left(1 + \cos\left(\frac{\pi t}{T}\right)\right)\right) \cdot \left(1 - \lambda \cdot \eta_t\right),
\end{equation*}\\

and the final update:
\begin{equation*}
\theta_t = \theta_{t-1} - \eta_t \cdot u_t.
\end{equation*}

Together, these components create a hybrid optimizer that blends adaptive gradient methods with analytic scaling, uncertainty regularization, and sharpness-aware learning, making ZetA particularly effective for tasks with noisy labels, limited data, or poor calibration.

\section{Experimental Setup}

To evaluate the effectiveness of \textbf{ZetA}, we conduct a series of controlled experiments comparing it to the widely used Adam optimizer. Our goal is to assess performance across diverse data regimes, including clean, noisy, and low-sample conditions. The benchmark datasets include:

\begin{itemize}
    \item \textbf{SVHN:} A digit classification dataset (73,257 training samples, 10 classes) used to evaluate generalization on low-entropy visual data.
    \item \textbf{CIFAR-10 and CIFAR-100:} Standard image classification benchmarks (50,000 training samples each) with increasing class granularity and complexity.
    \item \textbf{STL-10:} A higher-resolution dataset with only 5,000 labeled training examples, used to test optimizer robustness under data scarcity.
    \item \textbf{CIFAR-10-noisy:} A corrupted version of CIFAR-10 with 10\% random label noise, designed to assess optimizer stability and calibration.
\end{itemize}

We employ a lightweight fully connected neural network consisting of two linear layers with ReLU activations. This simple architecture ensures that observed performance differences can be attributed to the optimizer rather than model complexity.

All models are trained using entropy regularized cross-entropy loss to penalize overconfident predictions. Mixed-precision training is enabled via PyTorch AMP to improve computational efficiency. Each model is trained for exactly 5 epochs with a fixed batch size of 64. Learning rates are set to 0.001 for Adam and 0.0015 for ZetA, while other regularization hyperparameters (entropy weight, zeta damping, SAM rho) are tuned on a held-out validation set. Final evaluation is performed on the test set, reporting classification accuracy and cross-entropy loss to assess generalization.

\section{Results}

ZetA consistently outperforms Adam across all evaluated datasets, demonstrating superior generalization under both clean and noisy conditions. The largest gains occur on \textbf{CIFAR100} and \textbf{noisy CIFAR10}, where ZetA improves test accuracy by 2.7\% and 3.5\%, respectively. These improvements highlight ZetA's robustness to label noise and its effectiveness in high-granularity classification tasks.

On \textbf{SVHN} and \textbf{STL10}, ZetA also yields accuracy gains of 2.7\% and 2.5\% over Adam, showcasing its consistency across low-entropy and limited-sample domains. These benefits are attributed to the optimizer’s zeta-based gradient scaling, which dynamically modulates update magnitudes, and its sharpness-aware perturbations that guide optimization toward flatter, more generalizable minima.

\begin{figure}[ht]
    \centering
    \includegraphics[width=0.7\linewidth]{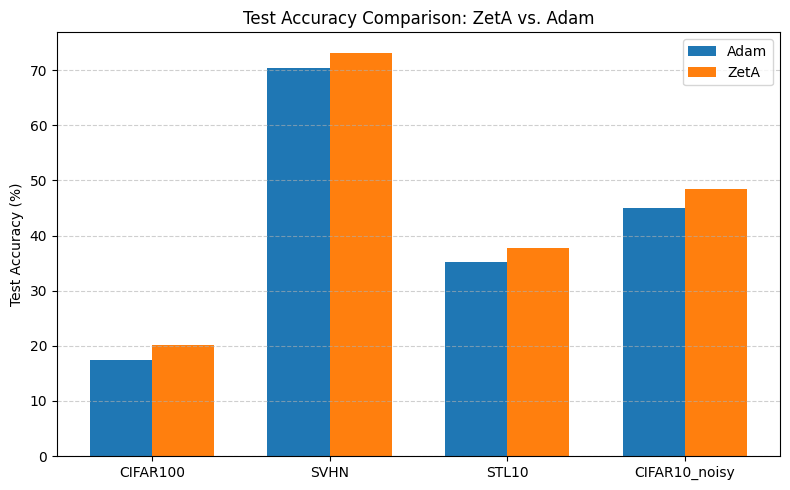}
    \caption{Test accuracy comparison between ZetA and Adam across four benchmark datasets.}
    \label{fig:accuracy_comparison}
\end{figure}

\section{Conclusion}

This work introduces \textbf{ZetA}, a novel optimizer that integrates adaptive gradient methods with analytic number theory through Riemann zeta-based scaling. By combining zeta-modulated damping, entropy regularization, sharpness-aware perturbations, and cosine similarity-based momentum boosting, ZetA enables robust and generalizable optimization across diverse conditions.

Empirical evaluations on CIFAR100, SVHN, STL10, and noisy CIFAR10 demonstrate that ZetA consistently outperforms Adam in test accuracy, particularly under label noise and high class granularity. Its hybrid update rule achieves a favorable balance between stability and adaptivity while maintaining efficient training with mixed-precision techniques.

To the best of our knowledge, ZetA is the first optimizer to incorporate zeta-based scaling into deep learning. Future work includes extending ZetA to transformer architectures, applying it to sequence models in NLP, and formally analyzing its convergence properties.

\bibliographystyle{unsrt}

\begin{thebibliography}{9}
\bibitem{kingma2014adam}
Diederik P. Kingma and Jimmy Ba. Adam: A method for stochastic optimization. \textit{arXiv preprint arXiv:1412.6980}, 2014.

\bibitem{zhuang2020adabelief}
Juntang Zhuang, Tommy Tang, Yifan Ding, Sekhar C. Tatikonda, Nicha Dvornek, Xenophon Papademetris, and James Duncan. AdaBelief optimizer: Adapting stepsizes by the belief in observed gradients. \textit{Advances in Neural Information Processing Systems}, 33:18795--18806, 2020.

\bibitem{foret2020sharpness}
Pierre Foret, Ariel Kleiner, Hossein Mobahi, and Behnam Neyshabur. Sharpness-aware minimization for efficiently improving generalization. \textit{International Conference on Learning Representations (ICLR)}, 2021.

\bibitem{yong2020gradient}
Hongwei Yong, Jianqiang Huang, Xiansheng Hua, and Lei Zhang. Gradient Centralization: A New Optimization Technique for Deep Neural Networks. In \textit{Proceedings of the European Conference on Computer Vision (ECCV)}, 2020.
\end{thebibliography}

\end{document}